\documentclass[twoside,11pt]{article}

\usepackage[abbrvbib, preprint]{jmlr2e}

\usepackage{jmlr2e}

\usepackage{graphicx}
\usepackage{amsmath,amssymb}
\usepackage{multirow}
\usepackage{subfigure}
\usepackage{comment}
\usepackage{wrapfig}
\usepackage{colortbl}
\usepackage{booktabs} 

\usepackage{listings}
\usepackage{color}

\definecolor{dkgreen}{rgb}{0,0.6,0}
\definecolor{gray}{rgb}{0.5,0.5,0.5}
\definecolor{mauve}{rgb}{0.58,0,0.82}

\usepackage{listings}
\usepackage{xcolor}

\definecolor{codegreen}{rgb}{0,0.6,0}
\definecolor{codegray}{rgb}{0.5,0.5,0.5}
\definecolor{codepurple}{rgb}{0.58,0,0.82}
\definecolor{backcolour}{rgb}{0.95,0.95,0.95}

\lstdefinestyle{mystyle}{
    language=Python,
    aboveskip=3mm,
    belowskip=3mm,
    backgroundcolor=\color{backcolour},   
    commentstyle=\color{codegreen},
    keywordstyle=\color{magenta},
    numberstyle=\tiny\color{codegray},
    stringstyle=\color{codepurple},
    basicstyle=\ttfamily\small,
    breakatwhitespace=true,  
    showstringspaces=false,
    breaklines=true,                 
    captionpos=b,                    
    keepspaces=true,                 
    numbers=left,                    
    numbersep=5pt,                  
    showspaces=false,                
    showstringspaces=false,
    showtabs=false,                  
    tabsize=3
}

\ShortHeadings{AlphaRotate: A Rotation Detection Benchmark using TensorFlow}{Xue Yang, Yue Zhou and Junchi Yan}
\firstpageno{1}

\begin{document}

\title{AlphaRotate: A Rotation Detection Benchmark using TensorFlow}

\author{\name Xue Yang\textsuperscript{1} \email yangxue-2019-sjtu@sjtu.edu.cn \\
      \name Yue Zhou\textsuperscript{2} \email sjtu\_zy@sjtu.edu.cn \\
       \name Junchi Yan\textsuperscript{1,}\thanks{Junchi Yan is the correspondence author.} \email yanjunchi@sjtu.edu.cn \\
       \addr \textsuperscript{1}Department of Computer Science and Engineering, MoE Key Lab of Artificial Intelligence, AI Institute, Shanghai Jiao Tong University, Shanghai, China, 200240\\
       \textsuperscript{2}Department of Electronic Engineering, Shanghai Jiao Tong University, Shanghai, China, 200240}

\editor{Kevin Murphy and Bernhard Sch{\"o}lkopf}

\maketitle

\begin{abstract}
AlphaRotate is an open-source Tensorflow benchmark for performing scalable rotation detection on various datasets. It currently provides more than 18 popular rotation detection models under a single, well-documented API designed for use by both practitioners and researchers. AlphaRotate regards high performance, robustness, sustainability and scalability as the core concept of design, and all models are covered by unit testing, continuous integration, code coverage, maintainability checks, and visual monitoring and analysis. AlphaRotate can be installed from PyPI and is released under the Apache-2.0 License. Source code is available
at \url{https://github.com/yangxue0827/RotationDetection}.
\end{abstract}


\section{Introduction}
Despite the rich literature~\citep{girshick2015fast,ren2015faster,dai2016r,lin2017feature,lin2017focal} in visual object detection in computer vision,  existing models are mostly agnostic to the object orientation, which only output a horizontal bounding box. This can be restricted in real-world settings whereby either the rotation information itself can be critical e.g. for aerial observation, or the rotated detecting bounding box can better align the object for more accurate recognition, especially for small and densely arranged objects.

For the above reasons, recently rotation detectors emerge with the development in terms of both backbone and loss design. The applications are arranged across aerial images~\citep{yang2018automatic, yang2018position, yang2019scrdet, yang2021r3det}, scene text~\citep{zhou2017east, jiang2017r2cnn, liu2018fots, ma2018arbitrary, liao2018rotation}, faces~\citep{shi2018real}, 3D objects~\citep{zheng2020rotation}, and retail scenes~\citep{chen2020piou, pan2020dynamic} etc.

However, there lacks an open-source benchmark integrating recent advance rotation detection models for evaluation and use. The most popular object detection benchmarks, e.g. MMDetection \citep{chen2019mmdetection}, Detectron2 \citep{wu2019detectron2}, SimpleDet \citep{chen2019simpledet}, are all focused on horizontal detection. AerialDetection\footnote{\url{https://github.com/dingjiansw101/AerialDetection}} is an earlier rotation detection benchmark based on MMDetection. However, it only provides some baselines and few methods, and it lacks maintenance and integration of new methods. Moreover, all these benchmarks are based on Pytorch~\citep{paszke2017automatic} which can be less efficient than Tensorflow~\citep{abadi2016tensorflow} in terms of industrial deployment.

\begin{figure}[!tb]
	\centering
	\subfigure{
		\begin{minipage}[t]{1\linewidth}
			\centering
			\includegraphics[width=0.98\linewidth]{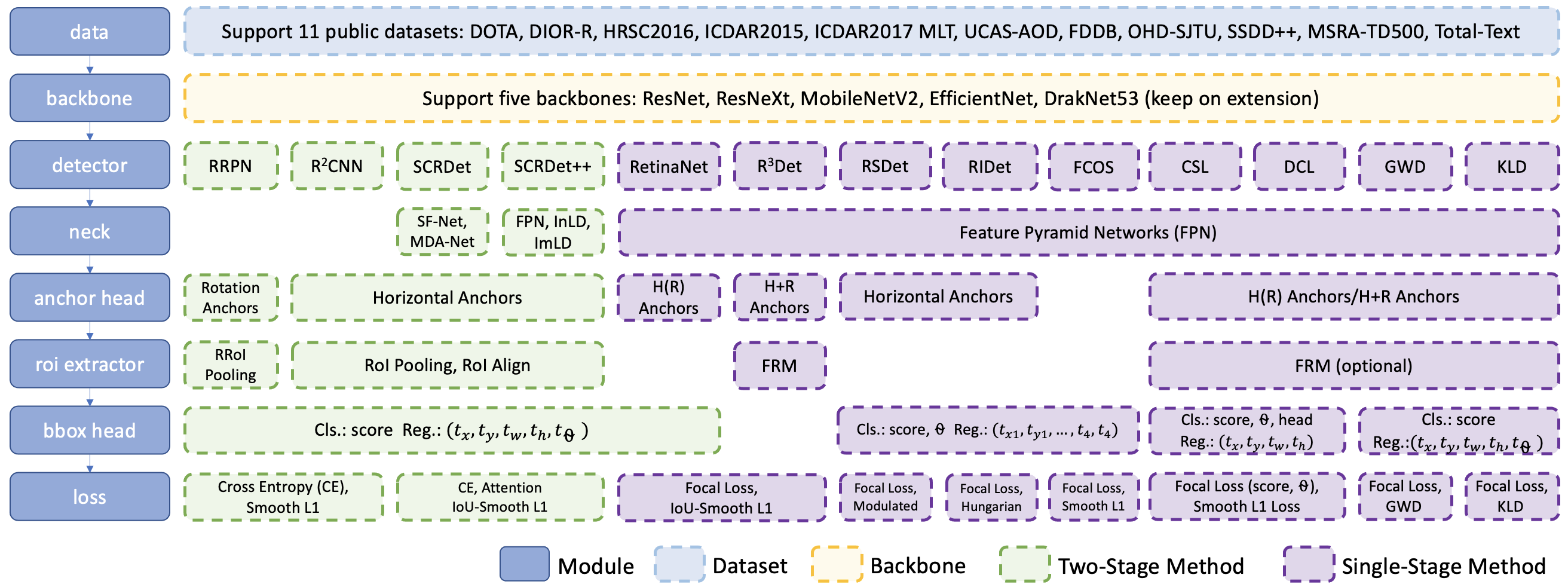}
		\end{minipage}
		\label{fig:h2r}
	} 
	\centering
	\label{fig:AlphaRotate}
	\vspace{-20pt}
	\caption{AlphaRotate supports state-of-the-art rotation detection methods, and supports training/testing on public datasets like aerial images, scene text, and face etc.}
	\vspace{-12pt}
\end{figure}

To fill this gap, we propose and implement AlphaRotate -- a Tensorflow based framework which consists of state-of-the-art detection techniques and models (keep on extension) -- mostly from our previous works in recent literature~\citep{yang2018automatic, yang2018position, yang2019scrdet,yang2020arbitrary,qian2021learning, yang2021dense,yang2021r3det,yang2021rethinking,yang2021learning}. It is user-friendly to both industry and academic community, with the following features, as shown in Figure~\ref{fig:AlphaRotate} and Table~\ref{tab:ablation_study}.

\begin{table}[tb!]
    \centering
    \resizebox{0.95\textwidth}{!}{
    \begin{tabular}{c|c|c|c|ccc|c|c}
        \toprule
        \multirow{2}{*}{Baseline} & \multirow{2}{*}{Improvement modules adding} & \multirow{2}{*}{Box Def.} & \multicolumn{1}{c|}{v1.0} & \multicolumn{3}{c|}{v1.0 train/val} & \multicolumn{1}{c|}{v1.5} & \multicolumn{1}{c}{v2.0} \\
        \cline{4-9} 
        & & & mAP$_{50}$ & mAP$_{50}$ & mAP$_{75}$ & mAP$_{50:95}$ & mAP$_{50}$ & mAP$_{50}$\\
        \hline
        \multirow{10}{*}{\shortstack{RetinaNet-H\\~\citep{lin2017focal}}} & - & $O.C.$ & 65.73 & 64.70 & 32.31 & 34.50 & 58.87 & 44.16 \\
        & - & $L.E.$ & 64.17 & 62.21 & 26.06 & 31.49 & 56.10 & 43.06 \\
        & IoU-Smooth L1~\citep{yang2019scrdet} & $O.C.$ & 66.99 & 64.61 & 34.17 & 36.23 & 59.16 & 46.31 \\
        & RSDet~\citep{qian2021learning} & $O.C.$ & 66.05 & 63.50 & 33.32 & 34.61 & 57.75 & 45.17\\
        & RSDet~\citep{qian2021learning} & Quad. & 67.20 & 65.15 & 40.59 & 39.12 & 61.42 & 46.71\\
        & RIDet~\citep{ming2021optimization} & Quad. & 66.06 & 64.07 & 40.98 & 39.05 & 58.91 & 45.35\\
        & CSL~\citep{yang2020arbitrary} & $L.E.$ & 67.38 & 64.40 & 32.58 & 35.04 & 58.55 & 43.34 \\
        & DCL (BCL)~\citep{yang2021dense} & $L.E.$ & 67.39 & 65.93 & 35.66 & 36.71 & 59.38 & 45.46 \\
        & GWD~\citep{yang2021rethinking} & $O.C.$ & 68.93 & 65.44 & 38.68 & 38.71 & 60.03 & 46.65\\
        & KLD~\citep{yang2021learning} & $O.C.$ & 71.28 & 68.14 & 44.48 & 42.15 & 62.50 & 47.69\\
        \hline
        \multirow{1}{*}{\shortstack{RetinaNet-R}} & - & $O.C.$ & 67.25 & 65.00 & 33.68 & 35.16 & 56.50 & 42.04\\
        \hline
        \multirow{2}{*}{\shortstack{FCOS\\~\citep{tian2019fcos}}} & - & Quad. & 67.69 & 65.73 & 35.70 & 36.62 & 61.05 & 48.10 \\
        & RSDet~\citep{qian2021learning} & Quad. & 67.91 & 66.07 & 38.90 & 38.25 & 62.18 & 48.81 \\
        \hline
        \multirow{4}{*}{\shortstack{R$^3$Det\\~\citep{yang2021r3det}}} & - & $O.C.$ & 70.66 & 67.18 & 38.41 & 38.46 & 62.91 & 48.43 \\
        & DCL (BCL)~\citep{yang2021dense} & $L.E.$  & 71.21 & 67.45 & 35.44 & 37.54 & 61.98 & 48.71 \\
        & GWD~\citep{yang2021rethinking} & $O.C.$  & 71.56 & 69.28 & 43.35 & 41.56 & 63.22 & 49.25 \\
        & KLD~\citep{yang2021learning} & $O.C.$ & 71.73 & 68.87 & 44.48 & 42.11 & 65.18 & 50.90 \\
        \hline
        \multirow{2}{*}{\shortstack{FPN\\~\citep{lin2017feature}}} & R$^2$CNN~\citep{jiang2017r2cnn} & $O.C.$ & 72.27 & 68.43 & 34.74 & 37.08 & 66.45 & 52.35\\
        & KLD~\citep{yang2021learning} & $O.C.$ & 72.16 & 68.78 & 39.25 & 39.58 & 65.59 & 51.30\\
        \bottomrule
    \end{tabular}}
    \caption{Accuracy comparison of rotation detectors on DOTA (v1.0, v1.5 and v2.0)~\citep{xia2018dota}. ${O.C.}$,  ${L.E.}$, and Quad. respectively denotes OpenCV Definition $\left(\theta \in [-90^\circ, 0^\circ)\right)$, Long Edge Definition $\left(\theta \in [-90^\circ, 90^\circ)\right)$, and quadrilateral box. `H' and `R' represent the horizontal and rotating anchors, respectively. All models are initialized by ResNet50~\citep{he2016deep} without using data augmentation and multi-scale training and testing. mAP$_{50}$ represents mean of Average Precision calculated under the IoU threshold of 0.5 (like-wise for mAP$_{75}$ and mAP$_{50:95}$).}
    \vspace{-20pt}
    \label{tab:ablation_study}
\end{table}

     \textbf{1)} AlphaRotate is one of the first rotation detection benchmarks (based on Tensorflow), and it supports training and testing on various datasets including aerial images, scene text, and face. It supports state-of-the-art rotation detectors (including hybrid methods), and provides comprehensive evaluation on DOTA~\citep{xia2018dota} dataset.
     
     \textbf{2)} A clean and modular implementation, which eases future integration of new methods and facilitates assembling different modules, such as Backbone, Neck, Loss, etc.
     
     \textbf{3)} AlphaRotate includes a detailed installation and tutorials across all models for clarity and ease of use, and all models are covered by unit testing, continuous integration, code coverage, maintainability checks, and visual monitoring and analysis.
     
     \textbf{4)} AlphaRotate supports multi-GPU training and multi-process testing, and provides commonly used techniques such as data augmentation, multi-scale training and cropping, stochastic weights averaging, etc. to further boost model performance.

\section{Project Focus}
\textbf{Modular implementation:} As shown in Figure~\ref{fig:AlphaRotate}, the detectors are organized by eight components: \emph{data}, \emph{backbone}, \emph{detector}, \emph{neck}, \emph{anchor head}, \emph{roi extractor}, \emph{bbox head}, \emph{loss}. The core of AlphaRotate is a set of base classes and functions that is designed to allow for rapid and easy development of such models. Developers only need to add/delete/improve specific modules to build a new rotation detection model. The modular design improves code utilization and also helps developers to debug and troubleshoot. Through our test procedure, all functions and classes are tested, with a line coverage of over 92\% of the code.

\noindent \textbf{Rich models and tools:} AlphaRotate supports more than 18 state-of-the-art rotation detection methods (keep on extension), including single-stage/two-stage methods, anchor-based/anchor-free methods, and supports training and testing on nearly ten datasets such as aerial images, scene text, and face, as shown in Figure~\ref{fig:AlphaRotate}. Besides, it provides fair comparisons of all methods (including hybrid methods) on representative DOTA dataset~\citep{xia2018dota} to provide researchers with an accurate and comprehensive baseline, as listed in Table~\ref{tab:ablation_study}. Unless otherwise specified, all models are trained on the trainval set by default, and the result file of the test set is submitted to the official evaluation server\footnote{\url{https://captain-whu.github.io/DOTA/evaluation.html}} to obtain the final evaluation result. We have provided download links of all baseline model weights.

\noindent \textbf{Open and collaborative development:} We have implemented AlphaRotate with an open and collaborativeh spirit and placed it under Apache-2.0 License. AlphaRotate is hosted on GitHub\footnote{\url{https://github.com/yangxue0827/RotationDetection}}, and developers can consult and discuss issues through the platform. In addition, external contributions and requests are encouraged and we enforce a strict rule on providing several tests for every new contribution and detected bug. When writing this paper, about 620 stars and 112 forks have been created and 47 issues have been resolved.

\noindent \textbf{Installation and tutorials:} AlphaRotate provides installation instructions, including dependent libraries and platform environment. We also provide developers with the Docker image matched by the benchmark. We also provide detailed tutorials\footnote{\url{https://alpharotate-tutorial.readthedocs.io/}}, including how to convert dataset, compile files, train and test, and use visualization tools, etc.

\noindent \textbf{Dependencies:} AlphaRotate is built as a Python application on Tensorflow~\citep{abadi2016tensorflow}. It allows to use the benchmark on various platforms and devices like CPUs and GPUs. AlphaRotate relies on open source libraries such as \emph{NumPy}, \emph{SciPy}, \emph{Cython}, \emph{OpenCV-Python}, \emph{Matplotlib}, \emph{Shapely}, which are commonly used in literature and community.

\section{Toolbox Usage}

\noindent \textbf{Data pre-processing:} The input of the model mainly includes the image $I$, the four-point coordinates of the object $(x_{1},y_{1},..,x_{4},y_{4})$ and the category corresponding to the object $L$. 
Then, dataset should be converted into tensorflow's exclusive data reading format, $tfrecord$.

\noindent \textbf{Model training:} First, select the detector and dataset, and create the corresponding configuration file according to the plan under $configs$. Then run the training script under $tools$ until the training is complete. During the training process, tensorbord can be used to observe the training status, including the loss curve of each stage and the image visualization. All trained models and log files are saved in the $output$. In the following example, the training process is easily implemented in a couple lines of code:
\begin{lstlisting}[language=python]
import alpharotate
from alpharotate.libs.models.detectors import retinanet
from retinanet.build_whole_network import DetectionNetworkRetinaNet
from configs import cfgs
# init detector (e.g. RetinaNet)
model = DetectionNetworkRetinaNet(cfgs=cfgs, is_training=True) 
# training
_, loss = model.build_whole_detection_network(input_img_batch=img, gt=gt) 
\end{lstlisting}

\noindent \textbf{Model testing and evaluation:} Under the $tools$, it provides test and evaluation scripts for evaluation (e.g. mAP, F-measure), visualizing results, and generating result files. 
\begin{lstlisting}[language=python]
# init detector (e.g. RetinaNet)
model = DetectionNetworkRetinaNet(cfgs=cfgs, is_training=False) 
# inference
bbox, score, category = model.build_whole_detection_network(input_img_batch=img) 
\end{lstlisting}

\section{Conclusion}
With the practical importance and academic emergence for visual rotation detection, AlphaRotate is a deep learning benchmark for visual object rotation detection in Tensorflow under the Apache-2.0 license. The architecture is designated for both flexibility and ease of use, with the goal of facilitating the deployment of rotation detection in diverse domains, both in industrial applications and in academic research. We will continue to improve the entire optimized benchmark and support representative detection methods in the future. We also welcome the community to participate in the development.



\vskip 0.2in
\bibliography{egbib}

\end{document}